# A Verifiable and Correct-by-Construction Controller for Robot Functional Levels[*]

Saddek Bensalem[†]  Lavindra de Silva[‡]  Félix Ingrand[‡]  Rongjie Yan[†]

*Abstract*—Autonomous robots are complex systems that require the interaction and cooperation between numerous heterogeneous software components. In recent times, robots are being increasingly used for complex and safety-critical tasks, such as exploring Mars and assisting/replacing humans. Consequently, robots are becoming critical systems that must meet safety properties, in particular, logical, temporal and real-time constraints. To this end, we present an evolution of the LAAS architecture for autonomous systems, in particular its G$^{en}_o$M tool. This evolution relies on the BIP component-based design framework, which has been successfully used in other domains such as embedded systems. We show how we integrate BIP into our existing methodology for developing the lowest (functional) level of robots. Particularly, we discuss the componentization of the functional level, the synthesis of an execution controller for it, and how we verify whether the resulting functional level conforms to properties such as deadlock-freedom. We also show through experimentation that the verification is feasible and usable for complex, real world robotic systems, and that the BIP-based functional levels resulting from our new methodology are, despite an overhead during execution, still practical on real world robotic platforms. Our approach has been fully implemented in the LAAS architecture, and the implementation has been used in several experiments on a real robot.

## I. Introduction

As robots are deployed for increasingly complex tasks, the need increases for proving that these systems are safe, dependable, and correct. This is particularly true for rovers used in expensive and distant missions, such as Mars rovers, that need to avoid equipment damage and minimize resource usage, but also for robots that have to interact regularly and in close contact with humans or other robots. Consequently, it may soon become common to require software integrators and developers to provide guarantees and formal proofs to certification bodies that, for instance, a rover will not move while it is communicating or even worse, while it is drilling, that the navigation software has no fatal deadlock, or that the arm of a service robot will not open its gripper while holding an object.

A certain level of dependability and safety can be achieved with thorough software testing and extensive simulation [1]. The goal of software testing is to "validate" and "verify" that the software meets a given set of requirements, and the goal of simulation is to detect errors as early as possible in the design phase. Unfortunately, both simulation and testing have the disadvantage of being incomplete, in the sense that each simulation run and each test evaluates the system only against a small subset of the foreseeable set of operating conditions and inputs. Hence, with complex autonomous and embedded systems it is often impractical to use these techniques to cover even a small fraction of the total operating space, not to mention the high cost of building test harnesses.

In this paper, we make a significant step toward building safe and dependable robotic systems. Robotic architectures (e.g., the LAAS architecture [2]) are typically organized into several levels, which usually correspond to different temporal requirements or different levels of abstraction of functionality. The lowest level of such architectures is the *functional* level, which includes all the basic, built-in action and perception capabilities such as image processing, obstacle avoidance, and motion control. We propose a novel approach for developing safe and dependable functional levels of complex, real-world robots. With our approach one can provide guarantees that the robot will not perform actions that may lead to states that are deemed unsafe, possibly resulting in undesired or catastrophic consequences.

Our solution relies on the integration of two state-of-the-art technologies, namely:

- G$^{en}_o$M [2] – a tool (part of the LAAS architecture toolbox) that is used for specifying and implementing the functional level of robots; and
- BIP [3] – a software framework for formally modeling complex, real-time component-based systems, with supporting toolsets for, among other things, verifying such systems, to guarantee *correctness-by-construction* (i.e. that the system is guaranteed correct from the rigorous design to implementation) to avoid as much as possible a posteriori verification.

This integration allows us to synthesize for our robots (e.g. Dala, an iRobot ATRV) a complete functional level that is correct by construction. "correct by construction" in this context, encompass the fact that all the basic specifications of a G$^{en}_o$M module (e.g., requests that are incompatible or requests for initialization) as well as the user-supplied (intra-module or inter-module) specifications and safety constraints will be enforced and formally guaranteed *online* by the underlying BIP model and the resulting controller. With the inclusion of such constraints, one can guarantee that the functional level will not reach unsafe states, even if bugs exist in user-supplied programs at higher levels of abstraction (e.g., the decisional level). Moreover, the resulting BIP model can be checked *offline* for properties such as deadlock-freedom using verification tools and suites. Indeed, one needs to find the right balance between constraining the BIP model to enforce properties on line, yet making sure it is deadlock-free to be able to run the real system live.

[*]Authors are in alphabetical order by last name. Part of this work is funded by the ESA/ESTEC GOAC project and by the FNRAE MARAE project.
[†]Verimag/CNRS, Grenoble I Uni., France. first.last@imag.fr
[‡]LAAS/CNRS, Toulouse Uni., France. first.last@laas.fr

*A. State of the art*

There are numerous works that address the architectural aspect of robot software development (e.g., CLARATY [4]), and software tools (e.g., OROCOS [5], CARMEN [6], Player Stage [7], Microsoft Robotics Studio [8], and ROS [9]) to develop the functional level of robotic systems. There is even some work which compares them [10], [11]. Yet none of these architectural tools and systems proposes any extension toward validation or verification. This is not to say that they could not be extended to allow it; for most of them this remains to be done.

Nevertheless, in the past there has been some attempts to incorporate some formal tools in the development of robotic platforms. Some of the requirements presented above were clearly addressed by a previous version of the LAAS architecture [12] based on the KHEOPS system [13]. KHEOPS is a tool for checking a set of propositional rules in real-time. A KHEOPS program is thus a set of production rules ($condition(s) \rightarrow action(s)$), from which a decision tree is built. More recently, we have proposed the $R^2C$ [14]. The main component of $R^2C$ is the *state checker*. This component encodes in an OBDD the constraints of the system, specified in a language named $Ex^oGen$. At run-time it continuously checks if new requests are consistent with the current execution state and the model of properties to enforce.

Another interesting early approach to prove various formal properties of robotic systems is the ORCCAD system [15]. This development environment, based on the Esterel [16] language, provides extensions to specify robot "tasks" and "procedures". However, this approach remains constrained by the synchronous systems paradigm.

In [17], the authors present another work related to synchronous languages, which has some similarities with the work presented here. Their objective is also to develop an execution control system with formal checking tools and a user-friendly language. This system makes use of an abstract representation of services (without explicit representation of arguments nor returned values). Their development environment allows the possibility to validate the resulting automata via model-checking techniques (with Sigali, a Signal extension).

In [18], the authors focus on code verification and certification using a theorem prover in high order logic. They illustrate their approach on the laser based navigation component of a mobile robot. As we will see, this work is complementary to ours, as it explicitly focuses on some aspects of verification (C/C++ code) which we do not handle. But their approach is somewhat limited to one particular component or algorithm, and does not manage the interactions between functional components.

In [19], the authors propose a system based on a model-based approach. The objective is to abstract the system into a state transitions based language abstracting the dependability concerns. The programmers specify state evolutions with invariants and a controller executes this maintaining these invariants. To do that the controller estimates the most likely current state—using observation and a probabilistic model of physical components—and finds the most dependable sequence of commands to reach specified goal (i.e., with a minimum failure probability).

In [20], the authors present the CIRCA SSP planner for hard real-time controllers. This planner synthesizes off-line controllers from a domain description (preconditions, post conditions and deadlines of tasks). CIRCA SSP can then deduce the corresponding timed automaton to control the system online, with respect to these constraints. This automaton can be formally validated with model checking techniques.

In [21], the authors present a system which allows the translation from MPL (Model-based Processing Language) and TDL (Task Description Language)—the executive language of the CLARAty architecture[4]— to SMV, a symbolic model checker language.

Compared to the approach we propose in this paper, these three latter works are mostly designed for the specification of the decisional level and do not address the validation and verification of the functional level.

More generally, as advocated in [22], an important trend in modern systems engineering is model-based design, which relies on the use of explicit models to describe development activities and their products. It aims at bridging the gap between application software and its implementation by allowing predictability and guidance through analysis of global models of the system under development. The first model-based approaches, such as those based on ADA, synchronous languages [23] and Matlab/Simulink, support very specific notions of components and composition. More recently, modeling languages, such as UML (Unified Modeling Language) [24] and AADL (Architecture Analysis and Design Language) [25], attempt to be more generic. They support notions of components that are independent from a particular programming language, and put emphasis on system architecture as a means to organize computation, communication, and implementation constraints. Component-based techniques for systems and software have not yet achieved a satisfactory level of maturity. Systems built by assembling together independently developed and delivered components often exhibit pathological behavior. Part of the problem is that developers of these systems do not have a precise way of expressing the behavior of components at their interfaces, where inconsistencies may occur. Components may be developed at different times and by different developers with, possibly, different uses in mind. Their different internal assumptions, when exposed to concurrent execution, can give rise to unexpected behavior, e.g., race conditions and deadlocks.

All these difficulties and weaknesses are amplified in embedded system design in general. They cannot be overcome, unless we solve the hard fundamental problems raised by the definition of rigorous frameworks for component-based design. Such kinds of frameworks require mathematical foundations that encompass both analytic and computation techniques. They should also possess the ability to cope with heterogeneity and constructivity during design, coming from the need to integrate components with different characteristics and to build blocks and glue components with known properties.

BIP differs significantly from existing component-based frameworks for software engineering. These often use multi-threaded programming and point-to-point interaction mechanisms, such as function calls, for coordination between components. In contrast, BIP executes atomic components concurrently and coordinates them in terms of high-level mechanisms such as protocols and scheduling policies.

Focusing on the organization of computation between components, BIP can be viewed as an architecture description language (ADL). Like other existing ADLs, such as Acme [26] and Darwin [27], BIP uses the concept of *connector* to express coordination between components. Nonetheless, connectors in BIP are stateless. The architecture, consisting of connectors and priorities, is clearly distinguished from behavior.

Another significant difference from other frameworks is that BIP is intended for system modeling. It directly encompasses timing and resource management. Other system modeling formalisms either seek generality to the detriment of rigorousness, such as SySML (Systems Modeling Language) [28] and AADL, or limit their scope to specific computation models, such as Ptolemy [29].

*B. Proposed approach*

Developing a functional level using our integrated framework combining the G$^{en}_o$M and BIP technologies consists of the following steps: *(i)* developing the functional level using the G$^{en}_o$M tool of the LAAS architecture; *(ii)* translating the G$^{en}_o$M functional level into an equivalent BIP model; *(iii)* adding safety constraints into the generated BIP model; and on one hand *(iv)* verifying offline the model with the D-Finder [30] tool in our BIP tool-chain, and on the other hand *(v)* using online the BIP model jointly with the BIP *engine* to run the robot.

Then, we can summarize the contributions of this paper as follows. First, we provide algorithms and data structures for generating from a given G$^{en}_o$M functional level specification an equivalent BIP functional level. We provide an implemented tool that can automate this translation process. Along with the BIP engine, the BIP functional level is used in place of its G$^{en}_o$M counterpart. Second, we show, using a Mars exploration scenario, how the user can straightforwardly use BIP to specify and enforce different kinds of safety constraints on a generated BIP functional level. Third, we present results from using D-Finder to incrementally verify the generated BIP functional level. In particular, we prove that a substantial part of the BIP functional level is deadlock-free, and we report, for the first time, experiences in using D-Finder (e.g., solutions to deadlocks encountered) with a complex, real-world domain.

This paper is organized as follows. In Section II, we present the existing LAAS architecture and the BIP tool-chain; in Section III, we discuss in detail how to generate from a G$^{en}_o$M functional level an equivalent BIP functional level; in Section IV, we show how the BIP engine can be used as a controller of the BIP functional level, in order to prevent the system from reaching "dangerous" states; in Section V, we show how D-Finder was used to analyze the BIP functional level for properties such as deadlocks; in Section VI, we give experimental results from using D-Finder for the verification of the BIP functional level, as well as a runtime performance comparison between the G$^{en}_o$M functional level and the BIP functional level; and finally, in Section VII, we present our conclusions and directions for future work.

## II. BACKGROUND

*A.* G$^{en}_o$M

As discussed in [31], the lowest level of most complex systems and robotic architectures is the *functional* level, which includes all the basic, built-in action and perception capabilities. These processing functions and control loops (e.g., image processing, obstacle avoidance, and motion control) are encapsulated into controllable, communicating modules. At LAAS, we use G$^{en}_o$M[1] [2] to develop these modules. Each module in the functional level of the LAAS architecture is responsible for a particular functionality of the robot. Complex functionality (such as navigation) is obtained by combining modules.

For example, the functional level of our Dala rover is shown in Figure 1. This functional level[2] includes two navigation modes. The first one, for mostly flat terrain, is laser based (`LaserRF`), and it builds a map (`Aspect`) and navigates using the near diagram (NDD) approach. In particular, *(i)* `LaserRF` acquires laser scans and stores them in the `Scan` poster, from which `Aspect` builds the obstacle map `Obs`; and *(ii)* `NDD` manages the navigation by avoiding these obstacles and periodically produces a speed reference (`speed`) to reach a given target from the current position `Pos` produced by `POM`. The speed reference produced by `NDD` is, in turn, used by `RFLEX`, which manages the low level robot wheels controller. `RFLEX` also produces the current position (`Pos`) of the robot based on odometry; this position is used by `POM` which manages the current position of the robot, as well as the various geometrical frame transformations of the robot sensors (e.g., Laser RF, cameras, and PTU) and effectors. The second navigation mode, for rough terrain, is vision based, and uses stereo images (`VIAM`, `Stereo` and `DTM`) to build a 3D map (`Env`), which is used as input into an arc based trajectory planner (`P3D`). `P3D` also produces a speed reference (`speed`) which can be used by `RFLEX`. `Hueblob`, using panoramic images taken by `VIAM`, monitors potentially interesting features in the images, and `PTU` controls the pan-and-tilt unit (PTU). Finally, `Antenna` emulates communication with an orbiter, and `Battery` and `Heating` emulate the management of the power/energy and temperature (respectively) of the whole platform.

All these modules are built by instantiating a generic template (Figure 2). Each module provides *services*, which can be invoked by the higher (decisional) level according to tasks that need to be achieved. Services can be *execution services*, which initiate *activities* that take time to execute, or *control services*, which take negligible time to execute and are responsible for setting and returning variable values[3]. For example, the NDD module provides five services corresponding to initializations of the navigation algorithm (*SetParams*, *SetDataSource*, and *SetSpeed*), and launching and stopping the computation of the motion direction toward a given goal (*Stop* and *GoTo*). Execution services are managed by *execution tasks*, responsible for launching and executing activities within the associated running services. Each module may "export" *posters* for others (modules or the decisional level) to read, which store data produced by the module. For example, the NDD module exports the Speed poster which is periodically read by the RFLEX module as the speed reference to apply on the wheels of the robot.

---

[1]G$^{en}_o$M and other tools from the LAAS architecture can be freely downloaded from: https://softs.laas.fr/openrobots/wiki/

[2]Module and poster names in Figure 1 are in the *fixed* font.

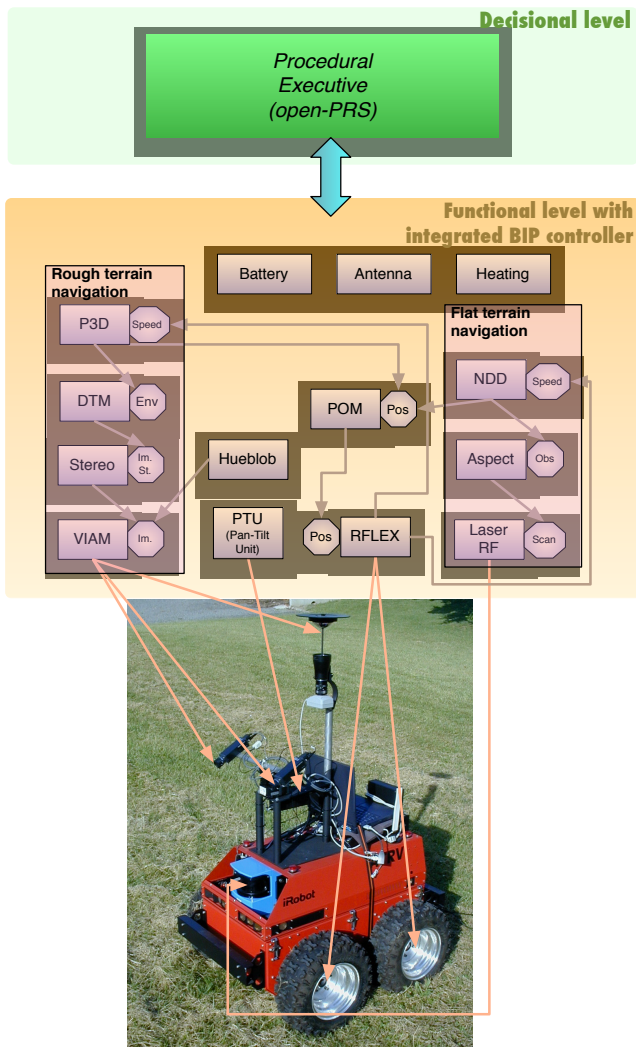

Fig. 1. The complete architecture of Dala.

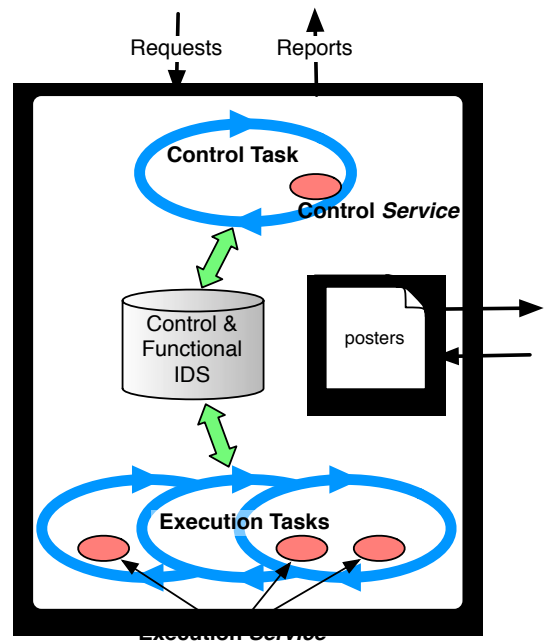

Fig. 2. A G$^{en}_o$M module internal organization.

Figure 3 presents the automaton of an activity. Transitions in the automaton correspond to the execution of particular elementary (C/C++) code, called *codels*, provided by the developers. Codels represent the core functionality of activities, and they are responsible for things such as initializing parameters (transition from *start* location), executing the "body" of the activity or its *main* codel (transition from *exec* location), and safely ending the activity, which may amount to things such as resetting parameters and sending error signals. As discussed in [2], the automaton in the figure depicts the following execution cycle. The activity is initially in the fictitious location *ether*. On receiving a *request(args)*, it transitions to the *start* location, during which parameters are checked and other running incompatible services interrupted. The activity is then started (location *exec*), from where it can keep executing (possibly looping back to this *exec* state if the execution spans over more than one cycle), be interrupted or ended. Finally, the activity returns an appropriate report and transitions back to location *ether*.

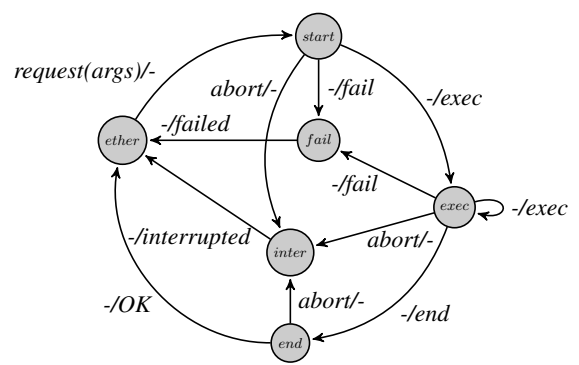

Fig. 3. The execution automaton of a G$^{en}_o$M activity. Transitions are of the form input/output.

---

[3] These variables are stored in a global data structure called *Functional-IDS*, which stores the data that is shared by all parts of a module. Similarly, there is a *Control-IDS* which stores some data about the internal control variables of a module.

*B.* BIP

BIP [3] is a framework[4] for modeling heterogeneous real-time programs.[5] The main characteristics of BIP are the following:

- BIP supports a model-based design methodology in which parallel programs are the superposition of three layers. The lowest layer specifies the behavior (*B*), the intermediate layer is a set of connectors specifying the interactions (*I*) between transitions of the behavior, and the upper layer is a set of priority rules (*P*) describing scheduling policies for interactions of the layer underneath. This layering offers a clear distinction between behavior and structure (i.e., connectors and priority rules).
- BIP uses a parameterized composition operator on programs, where the product of two programs is taken as the composition of their corresponding layers separately. Parameters are used to define the interactions as well as new priority rules between the parallel programs [32]. Such a composition operator facilitates incremental construction, i.e., obtaining a parallel program by successive composition of other programs.
- BIP provides a powerful mechanism for structuring interactions. This is done using *strong synchronization* and *weak synchronization*, where synchronous execution is characterized as a combination of the properties of the three layers.

The BIP framework is implemented in the form of a toolchain (Figure 4) which provides a complete implementation and a rich set of tools for things such as modeling, execution, and analysis (both static and on-the-fly). The tools can be broadly classified into:

- A front-end for editing and parsing BIP programs, and for generating an intermediate model. This can be used to generate code for execution and analysis on a back-end platform, as well as for static analysis.
- A back-end platform with an engine and the infrastructure for executing and analyzing the C++ application code, generated by the front-end.

*C.* BIP *language*

The BIP language [3], [33] supports the building of components from: *(i)* atomic components, which are a class of components in which interaction and priority layers are empty and behavior is specified as a set of transitions, and where triggers (labels) of transitions are ports (action names) used for synchronization; *(ii)* connectors, for specifying possible interaction structures between ports of atomic components; and *(iii)* priority rules, for selecting among possible interactions according to conditions, whose valuations depend on the state of the integrated atomic components.

An atomic component consists of: *(i)* a set of ports $P = \{p_1, \ldots, p_n\}$, where ports are used for synchronization with other components; *(ii)* a set of control states/locations $S = \{s_1, \ldots, s_k\}$, which denotes locations at which the components wait for synchronizations; *(iii)* a set of variables $V$ for storing (local) data; and *(iv)* a set of transitions encoding atomic computation steps.

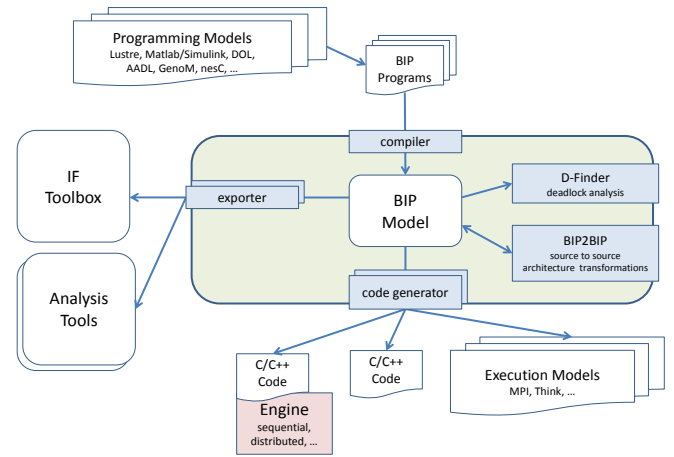

Fig. 4. The BIP tool-chain.

A transition is a tuple of the form $(s_1, p, g, f, s_2)$, representing a step from control location $s_1$ to $s_2$. A transition's execution is an atomic sequence of two microsteps: *(i)* an interaction including $p$, which involves synchronization between components, possibly with an exchange of data; followed by *(ii)* an internal computation specified by function $f$ on $V$. A transition can be executed only if the guard (boolean condition on $V$) $g$ holds and some interaction including port $p$ is offered.

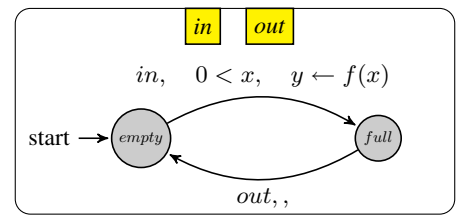

Fig. 5. A simple BIP atomic component.

Figure 5 shows a simple atomic component. This component has: two ports *in*, *out*; two variables $x$, $y$; and two control locations *empty*, *full*. At control location *empty*, a transition is possible if an interaction is possible on *in* and if $0 < x$. When the transition takes place, a new value for $y$ is computed. From control location *full*, a transition can occur when an interaction on *out* is possible. The omission of the guard and function for this transition means that the associated guard is true and the internal computation microstep is empty.[6]

An atomic component consists of a declaration of ports (identifiers) and data followed by the definition of its behavior. For data, basic C types can be used. In the behavior, the guard

---

[4]http://www-verimag.imag.fr/Rigorous-Design-of-Component-Based.html?lang=en

[5]Most material in this subsection and the next is based on [3].

[6]For legibility, we do not show in the rest of the paper, guards and functions in figures of BIP components.

and statement are respectively C expressions and statements. We assume that these are written with appropriate care to respect the atomicity assumption for transitions, e.g., no side effects and guaranteed termination. The behavior is defined by a set of transitions, where a transition is specified by a control location, followed by a port identifier, guard, function, and a target location. The BIP description of the reactive component of Figure 5 is the following:

**component** *Reactive*
  **port** *in,out*
  **data int** *x,y*
  **behavior initial to** *empty*
    **state** *empty*
      **on** *in* **provided** $0 < x$ **do** $y \leftarrow f(x)$ **to** *full*
    **state** *full*
      **on** *out* **to** *empty*
  **end**
**end**

Compound components are constructed from a set of atomic components with disjoint sets of names for ports, control locations, variables, and transitions. A set of ports that contains at most one port from each atomic component is called an interaction. We simplify the notation of an interaction by writing $p_1 p_2 p_3 p_4$ for the set of ports $\{p_1, p_2, p_3, p_4\}$. A connector, denoted by $\gamma$, is a set of interactions. For instance, if $p_1$, $p_2$, and $p_3$ are ports of distinct atomic components, then the connector $\gamma$ over $\{p_1, p_2, p_3\}$ may have seven interactions: $p_1$; $p_2$; $p_3$; $p_1 p_2$; $p_1 p_3$; $p_2 p_3$; and $p_1 p_2 p_3$. Given a connector $\gamma$, there are, as mentioned before, two basic modes of synchronization: *(i)* strong synchronization (or *rendezvous*), when the only feasible interaction of $\gamma$ is the maximal one, i.e., the one containing all the ports of $\gamma$ (i.e., $p_1 p_2 p_3$ in the example above); and *(ii)* weak synchronization (or *broadcast*), when the feasible interactions are all the interactions containing a particular active port which initiates the broadcast (i.e., if $p_1$ initiates the broadcast in the above example, the possible interactions are: $p_1$; $p_1 p_2$; $p_1 p_3$; and $p_1 p_2 p_3$).

In BIP syntax, a connector specification includes its set of ports, followed by an optional guarded list of its feasible interactions, each with an optional behavior. If the list is omitted, all feasible interactions with respect to the connector's set of ports can be unconditionally executed with no additional behavior execution. For instance, if there are no **on** statements specified in the broadcast connector below, all synchronizations involving appropriate combinations of ports mentioned in the **connector** statement can occur. The behavior specification of a connector, like that of a transition, is a set of guarded commands associated with feasible interactions. Specifically, if $\alpha = p_1 p_2 \ldots p_n$ is a feasible interaction, then its behavior is described by a statement of the form: **on** $\alpha$ **provided** $G_\alpha$ **do** $F_\alpha$, where $G_\alpha$ and $F_\alpha$ are respectively a guard and a statement representing a function on the variables of the components involved in the interaction. An example of the syntax of a connector is given below. Note that this syntax is a simplified version of the one given in the BIP literature.

**connector** $conn(c_1.p_1, c_2.p_2)$
**define** $[c_1.p_1{}', c_2.p_2]$
**on** $c_1.p_1, c_2.p_2$
**provided** $g$
**do** $\{c_2.p_2.v \leftarrow C\}$
**on** $c_1.p_1$
**provided** $g$
**do** $\{\}$

This connector, called *conn*, is a broadcast connector due to there being a port–$p_1$–that initiates the broadcast, denoted by the closed inverted comma next to $p_1$ in the **define** statement. Thus, port $p_1$ of component $c_1$ is the initiator of the broadcast synchronization between ports $c_1.p_1$ and $c_2.p_2$, where $c_2$ is a component and $p_2$ is one of its ports. If a strong synchronization involving both ports can occur, then a data transfer takes place, i.e., the variable $v$ of port $p_2$ is assigned the constant $C$. No synchronization can take place unless guard $g$ is met.

Finally, a compound component allows defining new components from existing sub-components (atoms or compounds) by creating their instances, and by specifying the connectors between them and the priorities between connectors. Component instances can have parameters for assigning initial values to their variables through a named association.

### D. D-Finder

The D-Finder tool[7] implements a compositional [30] and incremental methodology [34], [35] for the verification of component-based systems described in the BIP language [3]. D-Finder is mainly used to check safety properties of composite components, particularly useful for checking deadlock-freedom. To check safety properties, D-Finder applies the compositional verification method proposed in [30], [34], [35]. In this method, the set of reachable states is approximated by component invariants and interaction invariants. Component invariants are over-approximations of the set of the reachable states of atomic components and are generated by simple forward propagation techniques. Interaction invariants express global synchronization constraints between atomic components.

When we are concerned with the verification of global deadlock properties, we let $DIS$ be the set of global states in which a deadlock can occur. A deadlock is global, meaning that no interaction is enabled in the deadlock state. Therefore, $DIS$ can be computed by considering the disabled conditions of all the interactions. The tool progressively finds and eliminates potential deadlocks as follows. D-Finder starts with a BIP model as input and computes component invariants $CI$ by using the technique outlined in [30]. From the generated component invariants, D-Finder computes an abstraction of the BIP model and the corresponding interaction invariants $II$. From the enabled conditions of the interactions, D-Finder computes global deadlocks $DIS$. Then, it checks satisfiability of the conjunction $II \wedge CI \wedge DIS$. If the conjunction is unsatisfiable, then there is no deadlock. Otherwise, either D-Finder generates stronger component and interaction invariants or it tries to confirm the detected deadlocks by using reachability analysis techniques. When verifying other safety properties with D-Finder, one of the steps is to replace $DIS$ by the set of global states in which property violations occur. The other steps are identical to those of deadlock-freedom checking.

---
[7] http://www-verimag.imag.fr/DFinder.html

*E.* BIP *engine*

The BIP tool-chain provides a platform for executing and analyzing the C++ application code generated by the front-end. The tool-chain includes an engine and the associated software infrastructure. The engine, entirely implemented in C++ on Linux, directly implements the operational semantics of BIP.

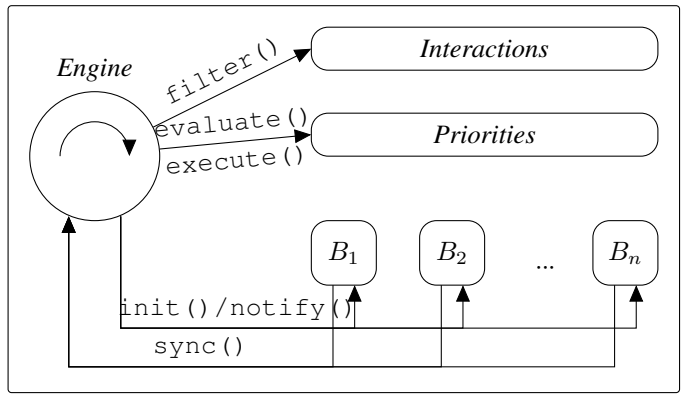

Fig. 6. The centralized engine architecture.

The engine works based upon the complete state information of the components. The execution follows a two-phase protocol, marked by the execution of the engine, and the execution of the atomic components. In the execution phase of the engine, it computes the interactions possible from the current state of the atomic components, and guards of the connectors. Then, between the enabled interactions, priority rules are applied to eliminate the ones with low priority. During this phase, the components are blocked, and await to be triggered by the engine. The engine selects a maximal enabled interaction, executes its data transfer, and triggers the execution of the atomic components associated with this interaction. The second phase is the execution of the local transitions of the notified atomic components. In our implementation, they continue their local computation independently in their own thread and eventually reach new control states. Here, the atomic components notify the engine regarding their enabled transitions and get blocked once more. The two phases are repeated, unless a deadlock is reached or the user wants to terminate the simulation. The scheme of the protocol is shown in Figure 6.

### III. COMPONENTIZATION OF THE G<sup>en</sup>oM FUNCTIONAL LEVEL

In this section, we discuss the componentization of the functional level, i.e., how the functional level is organized into a collection of separate modules, each with a specific functionality and a well-defined interface. In particular, we discuss our algorithms for mapping a given G<sup>en</sup>oM functional level into an equivalent BIP functional level. We start with the mapping from generic G<sup>en</sup>oM modules (Figure 2) to their BIP counterparts. The generic G<sup>en</sup>oM module is mapped to a hierarchy of BIP components, as shown in Figure 7. In addition to representing some G<sup>en</sup>oM entity, each box in this figure also represents an atomic or compound BIP component. In the same way we instantiate the generic G<sup>en</sup>oM module to produce a particular G<sup>en</sup>oM module, we can instantiate the generic BIP model (of the generic G<sup>en</sup>oM module) to obtain the BIP model of a particular module.

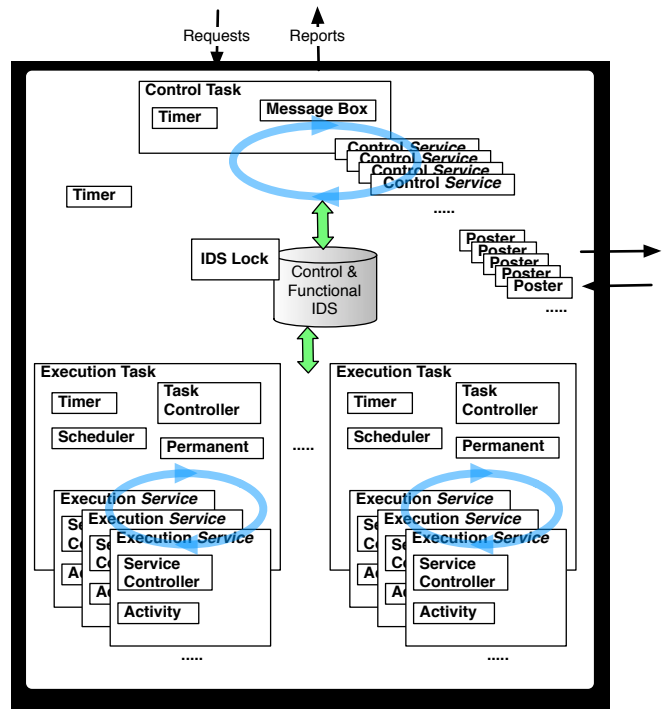

Fig. 7. A G<sup>en</sup>oM module BIP componentization.

In the componentization, an *Execution Task* is a compound component consisting of: a *Scheduler* (atomic) component, to control the execution of the *Activity* component of some *Execution Service* component; a *Task Controller* (atomic) component to stop the *Scheduler* if none of its associated *Execution Service* components are running; a *Timer* component to control the execution period of the *Execution Task*; and a *Permanent* component if the *Execution Task* has a permanent codel[8].

The *Poster* components are used for exporting the data stored within the module's Functional-IDS data structure, and they provide operations for reading from and writing to this data. The *IDS Lock* component represents a semaphore for ensuring mutual exclusion between different *Execution Task* components and *Execution Service* components when manipulating *Poster* components. The *Timer* component (directly) in the *Module* component is used by *Poster* components to determine how much time has elapsed, in terms of "ticks," since the last modification to their data. Specifically, *Poster* components contain a variable called *PosterAge* (initially 0) that is incremented for each "tick" activated by the associated *Timer* component, and reset whenever the poster is written to. The *Service Controller*, *Activity*, and *Control Task* components are discussed later.

---

[8]A permanent codel is executed at each loop of an execution task. It is not defined nor associated to a particular request.

As shown in Figure 7, some of the atomic components are combined to form compound components such as *Execution Service*. This is done by adding the necessary connectors between the atomic components. In turn, these compound components are combined using connectors to form the even more abstract component *Module*, corresponding to a G$^{en}$₀M module. By combining components incrementally (or "bottom-up") in this way, we have the guarantee that if its constituent components are proven to be correct (with respect to certain safety properties), then the resulting compound component is also correct, provided it is free of deadlocks [35]. The reason being that the already established safety properties are not violated by adding stronger interactions over the constituent components. Checking for deadlock-freedom using D-Finder is discussed in Section V.

The most important components from those in Figure 7 are *Message Box*, *Activity*, and *Service Controller*. Each *Module* component has, within its *Control Task* component, a *Message Box* component (Figure 8), which represents the interface for receiving requests for services and sending back replies. The period with which requests are read is controlled by the *Timer* component of the *Control Task*. There are two approaches for handling a newly received request in a *Message Box*: either *(i)* reject the request along with a specific report explaining the reason; or *(ii)* unconditionally accept the request. The latter is done via two transitions. The first transition ($abtInc_{b_i}$, where $b_i$ represents an *Execution Service* component) is for implementing a G$^{en}$₀M feature of interrupting certain execution services (*Execution Service* components) that are incompatible with the new request, and the second transition ($trig_{b_i}$) actually executes the request by interacting with either a *Control Service* or an *Execution Service* component.

Each *Execution Service* component has one corresponding *Service Controller* component (Figure 9), which controls its execution by, for example, checking the validity of the parameters (if any) of the request (e.g., *GoTo*) associated with the service, and handling the aborting of the service's execution. The *Service Controller* component has two variables: *active* and *done*, which are both initially *false*. The execution of the *Service Controller* starts via a synchronization with port *trig*, which sets *active* to *true*, after which the service can be aborted from any location via synchronization with the *abt* port. On the two transitions to location *ethr*, variable *active* is set to *false*, and variable *done* is set to *true* provided the transition labeled *fin* (denoting successful completion of the activity) was taken. Like a G$^{en}$₀M activity, the execution of the main codel of the *Service Controller* is initiated by the *exec* transition from the *exec* location. In each location of the *Service Controller* the status of the service can be obtained by synchronizing with the *stat* port of the component. Note that *Service Controller* components belonging to different *Module* components are equivalent except for the code executed during certain transitions (e.g., the ones labeled *ctrl* and *abt*), and that transitions labeled *abt* have higher priority than all other transitions, i.e., whenever a transition labeled *abt* and some other transition are both possible, the former is taken instead of the latter.[9]

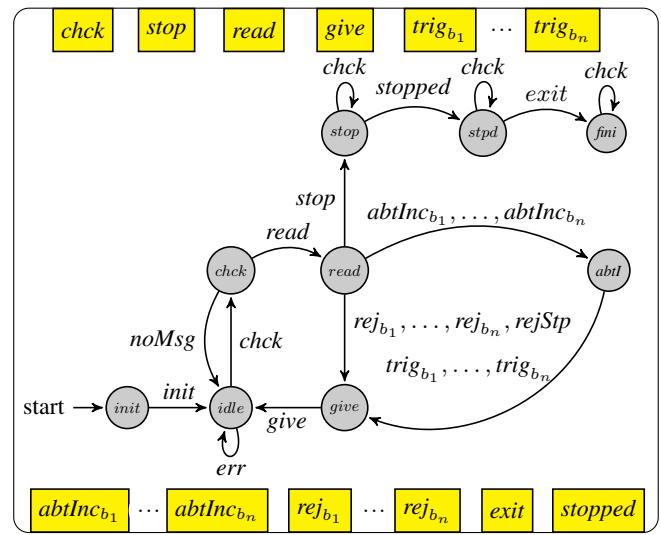

Fig. 8. An (atomic) BIP *Message Box* component. Transitions with multiple ports (separated by commas) represent multiple such transitions, each with one of the ports. Port *init* initializes the component; *chck* checks if there is a new request to service; *stop* stops the component from serving requests; and *err* "kills" the component with an error message if the initialization was unsuccessful.

The *Activity* component (not shown) corresponds to the G$^{en}$₀M activity automaton shown in Figure 3. This component waits to be initiated, i.e., for a synchronization between its *start* port and the *start* port of *Service Controller*, and then waits for its main execution to begin, i.e., for a synchronization between its *exec* port and the *exec* port of *Service Controller*. In particular, this latter synchronization leads, in the *Activity* component, to a transition that executes the main codel of the associated G$^{en}$₀M execution service. The *Activity* component is aborted by synchronizing with the *abt* port of the corresponding *Service Controller*; this synchronization sets a flag that prevents the *Activity* component from executing its main codel again.

It is worth noting that, unlike its G$^{en}$₀M counterpart, each BIP *Execution Service* component has exactly one *Service Controller* component, and consequently, a given *Execution Service* component cannot have more than one *Activity* component associated with it (and executing concurrently). While this is indeed limiting, it can be overcome to a certain extent by defining additional execution service types in the G$^{en}$₀M module (which yields additional *Execution Service* components in the corresponding BIP module) to cater for the additional activity instances that may be needed at runtime. For example, when simulating battery usage, there is a request called *StartUsingBattery*—which takes the device, e.g., camera, as an argument—to register the fact that the device in question has started to use the battery, and another request called *EndUsingBattery* to register the fact that the device has stopped using the battery. Since this design may result in multiple G$^{en}$oM activities at runtime for the execution service associated with the *StartUsingBattery* request (specifically, if multiple devices need to concurrently use the battery), we instead have multiple requests, such as *CameraStartUsingBattery*, one for each device that needs to use the battery.

---

[9]This is necessary to ensure that a request to abort a service is given priority over, for instance, starting another step in its execution.

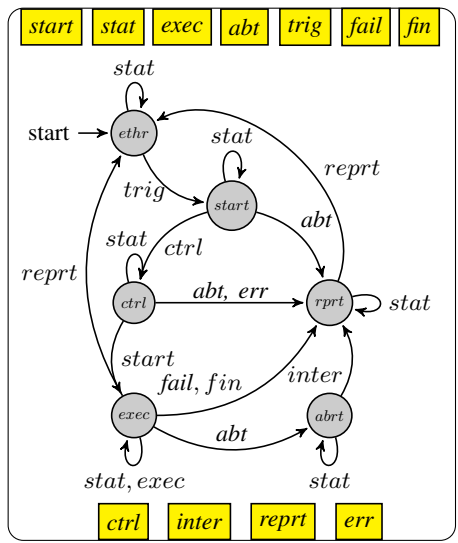

Fig. 9. An (atomic) BIP *Service Controller* component.

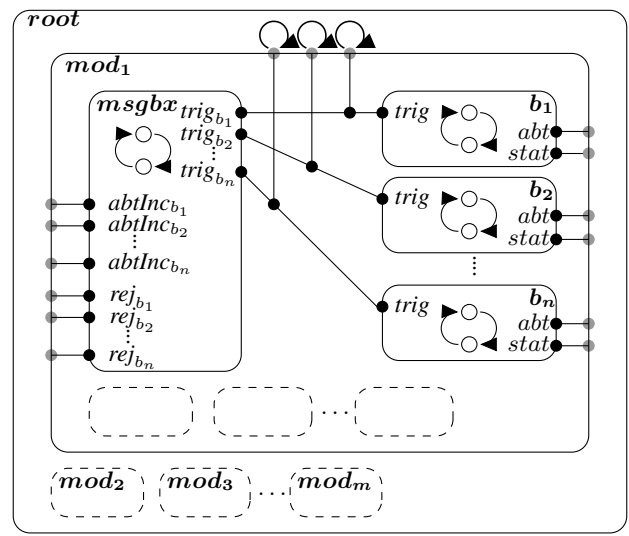

Fig. 10. A high-level illustration of exported ports (in grey), and important connectors within *Module* components ($mod_1, \ldots, mod_m$). Connectors are between a *Message Box* ($msgbx$) and the associated *Service Controller* components ($b_1, \ldots, b_n$).

Observe from Figure 8 that, due to the semantics of BIP components, an interaction involving a $trig_{b_i}$ port of a *Message Box* component cannot happen concurrently with an interaction involving some other $trig_{b_j}$ port of the component. Similarly, we have added the restriction that no interaction involving a $trig_{b_i}$ port of a *Message Box* component can happen concurrently with an interaction involving a $trig_{b_j}$ port of some other *Message Box* component (belonging to some other *Module* component). Without this restriction it would not be possible to add connectors (as we do later in Section IV) to guarantee that two services are not triggered concurrently.[10] To add this restriction, we use a simple atomic component which represents a semaphore; its token is obtained by the $read$ transition of *Message Box* components, and released by the $give$ transition of *Message Box* components. [11]

As shown in Figure 10, each port $trig_{b_i}$ of a *Message Box* component is synchronized via rendezvous with port $b_i.trig$ of *Service Controller* component $b_i$. All such connectors are exported so that they are "visible" from the root component, i.e., it is possible to interact with them from the root component. The root component in BIP is the top-level (compound) component that includes all the other components. In our case, the root component includes all components of the functional level.

From now on, for convenience, we simply use $trig_{b_i}$ to refer to such an exported connector involving a port $trig_{b_i}$. For example, the exported port (grey circle) in Figure 10 that is associated with the connector involving $trig_{b_1}$ is also referred to as $trig_{b_1}$. Similarly, for each *Service Controller* component $b_i$, the $rej_{b_i}$ and $abtInc_{b_i}$ ports of the associated *Message Box* component, as well as the $b_i.abt$ and $b_i.stat$ ports of $b_i$ are exported so that it is possible to interact with them from the root component. As before, from now on we simply use $rej_{b_i}$, $abtInc_{b_i}$, $b_i.abt$ and $b_i.stat$ to refer to these exported ports. Unlike the other ports shown in Figure 10 (such as $abt$), by default, an interaction on each $trig_{b_i}$ port is possible due to a singleton connector with no guard (i.e., a "no-op" connector) at the root-component level for each $trig_{b_i}$ port. On the other hand, by default, all $rej_{b_i}$, $abtInc_{b_i}$, $b_i.abt$ and $b_i.stat$ ports are not executable due to all such ports being left unconnected at the root-component level.

To ease the integration of BIP in the new framework, we have developed a tool that automatically produces a BIP model from a G$^{en}$oM module description file. Using this tool we have converted all the G$^{en}$oM modules in Figure 1 into their corresponding BIP modules, except for the POM module whose functionality is only remotely useful in this particular experiment. Instead of the original G$^{en}$oM modules, we are now using the resulting BIP modules within the top-level (root) BIP component representing the Dala functional level. Still, if one wants to enforce some safety properties inside a module (intra-module) or between modules (inter-module), these constraints have to be explicitly added into individual

---

[10] Note that this restriction does not imply that two arbitrary *Service Controller* components cannot be executed concurrently. That is, two *Service Controller* components can interact with their corresponding *Activity* components or abort the services at the same time when they have been triggered respectively by component *Message Box*.

[11] More specifically, the semaphore component has two locations; a $take$ transition from the initial location to the other location, and a $give$ transition to the initial location from the other location; moreover, for each *Message Box* component, a rendezvous connector exists between the *Message Box* component's $read$ transition and the $take$ transition of the semaphore component, and between the *Message Box* component's $give$ transition and the $give$ transition of the semaphore component.

BIP modules or into the root BIP component. We discuss such constraints in the next section.

## IV. FUNCTIONAL LEVEL CONTROLLER SYNTHESIS

Since commands to the functional level are sent from the decisional level (Procedural Reasoning System (OPENPRS) [36] in our case), and programs written for the decisional level may contain erroneous handcoded procedures, we need to constrain the functional level to ensure the appropriate/safe execution of $G^{en}_oM$ services. For example, despite what may be encoded in decisional level procedures, one may want to ensure that there is never a situation in which too much power is drawn from the battery, that the speed reference produced by a navigation mode is "fresh" enough with respect to the sensor data that it uses, or that the robot does not move when it is taking high resolution pictures or communicating.

In the previous LAAS architecture, the proper execution of $G^{en}_oM$ services was managed by a centralized controller called $R^2C$ [14], [37]. The purpose of such a controller is to prevent the system from reaching dangerous states, such as those mentioned, which could lead to undesirable or catastrophic consequences. The $R^2C$ controller maintains its own model and global state of the system. For the latter, $R^2C$ monitors all requests sent from the decisional level to the functional level and all reports sent back from the functional level. If a request sent to the functional level will lead the system into an undesirable (or unsafe) state, $R^2C$ takes actions to prevent the state from being reached, such as killing a service or rejecting the request. Otherwise, $R^2C$ allows the request to go through and the result to be returned.

The $R^2C$ *state checker* encodes the constraints of the system, specified in a language named EX$^O$GEN. This language includes a set of predicates such as *past(c)*, which checks if the last service instance successfully completed satisfies the constraint *c*, and predicate *in*, which checks if the value of a resource is a member of some given domain. An example of an EX$^O$GEN specification for the *GoTo* service of the NDD module is shown below [37]:

```
request GoTo(?goal) {
  fail {
    maintain:
    past(LocalizationSet(?mode) with ?mode==STEREO) &&
      (!past(StereoStart()) || [StereoStart() < StereoStop()]);
    past(LocalizationSet(?mode) with ?mode==GPS) &&
      GPSStatus in [0,0.1];
  }
}
```

In this specification, the *GoTo* service cannot run if the localization mode is *STEREO* and either *(i)* the stereo was never started, or *(ii)* the last time it was started precedes the last time it was stopped; or if the localization mode is *GPS* and the GPS reception level is below 10%. If at least one of these constraints become true then all running instances of *GoTo* are stopped. The EX$^O$GEN specifications are used to generate the state checker, which has a similar structure to OBDDs [38].

The state checker is used to maintain online the desired system state by, for example, ensuring the proper execution order of $G^{en}_oM$ services.

One of the main differences between the $R^2C$ approach and the BIP approach is that the former merely acts as a "filter" below the decisional level to enforce constraints between requests, while relying mostly on the control provided by $G^{en}_oM$. The BIP model and engine, on the other hand, go far beyond this by providing a formal and much finer grained model of the control taking place inside a functional module, which allows the user to specify finer grained constraints on the behavior of functional modules. Moreover, in our new framework, we have one integrated system with a single model and single global state, rather than two systems ($G^{en}_oM$ and $R^2C$) with two different models and two (possibly inconsistent) representations of the global state. Finally, by using BIP we now have a clearer semantics for constraints specified as BIP connectors, compared to the semantics of constraints specified in $R^2C$.

Before discussing how safety constraints can be encoded as BIP connectors, we first discuss the Mars Rover scenario we have written for the decisional level using the OPENPRS [36] executive. We use this scenario to motivate and illustrate our constraints. Our scenario consists of the rover heating up to a given temperature, and then exploring a predefined set of locations, which involves navigating to them and then taking science pictures at these particular locations. For now, we only use laser based navigation, via modules `LaserRF`, `NDD` and `Aspect`, as discussed in Section II-A. In the near future we plan to completely incorporate vision based navigation, via modules `P3D`, `DTM` and `Stereo`, into the BIP functional level and OPENPRS scenario, for example, by adding safety constraints between these modules and the ability to switch between the two navigation modes based on environmental factors, such as whether the robot is on flat terrain (laser based navigation) or rough terrain (vision based navigation). After navigating to a location, taking a science picture involves aligning the high resolution cameras—mounted on the PTU— to face the surfaces near the left and right front wheels of the rover. During navigation, the rover continuously monitors its surroundings for bright red rocks (objects) using the low resolution panoramic camera mounted on the mast (we call this opportunistic science). If such a rock is found, the rover stops navigating, determines if the rock is still within its front cameras' visibility area, and then takes a picture of the rock by aligning the front cameras toward it. The rover transmits all new images (both high and low resolution ones) to the orbiter during given visibility windows. Once all locations have been explored, the rover navigates back to its original location.

Now we can discuss in detail some of the constraints we have added into the BIP functional level (i.e., *root* component). We split the constraints into *intra-module* constraints, i.e., those between services belonging to a single module, and *inter-module* constraints, i.e., those between services belonging to two or more modules.

In what follows, ports with suffixes *Trigger*, *Reject*, *Abort*, *Status*, and *AbortIncompatibleServices* are used to represent (respectively) particular $trig_b$, $rej_b$, $b.abt$, $b.stat$, and $abtInc_b$ ports. Recall from Section III that all of these are exported ports.

## A. Intra-module constraints

In the NDD module, there must be at least one successfully[12] completed *SetParams* (SetParameters) service, and at least one successfully completed *SetSpeed* service before a *GoTo* service can be triggered. This constraint is to ensure that the robot does not start moving at an arbitrary (and potentially dangerous) speed. In what follows, *SPS = SetParamsStatus*, *SSS = SetSpeedStatus*, *GT = GoToTrigger*, and *GR = GoToReject*.

**connector** *AllowGoToIfArgsSet*(*ndd.GT*, *ndd.SPS*, *ndd.SSS*)
**define** [*ndd.GT*, *ndd.SPS*, *ndd.SSS*]
**on** *ndd.GT*, *ndd.SPS*, *ndd.SSS*
**provided** *ndd.SPS.done* $\wedge$ *ndd.SSS.done*
**do** {}

**connector** *RejectGoToIfArgsNotSet*(*ndd.GR*, *ndd.SPS*, *ndd.SSS*)
**define** [*ndd.GR*, *ndd.SPS*, *ndd.SSS*]
**on** *ndd.GR*, *ndd.SPS*, *ndd.SSS*
**provided** $\neg$*ndd.SPS.done* $\vee$ $\neg$*ndd.SSS.done*
**do** {*ndd.GR.rep* $\leftarrow$ PARAMS-OR-SPEED-NOT-SET}

Connector *AllowGotoIfArgSet* will trigger the *GoTo* service when both *SetParams* and *SetSpeed* services are successfully completed. Connector *RejectGotoIfArgsNotSet* will reject the request for the *GoTo* service along with an error message indicating that either the *SetParams* or *SetSpeed* service was not successfully completed.

## B. Inter-module constraints

This section discusses constraints involving multiple modules. First, pictures should not be taken with any high resolution camera while the robot is moving, and vice versa, in order to prevent high resolution pictures from being blurred (this constraint does not apply to low resolution panoramic pictures). Hence, we say that moving is "incompatible" with taking a picture with a high resolution camera. To enforce this constraint, whenever a new request is received that is incompatible with a currently executing service, we can either abort the latter with a specific error message and execute the new request, or reject the new request with a specific error message. Below we show an example of the case where the new request is rejected. In what follows, note that *AcS = AcquireStatus*, *TSST = TrackSpeedStartTrigger*, and *TSSR = TrackSpeedStartReject*. We only show the first two connectors; the other two (i.e., *AllowAcquireIfNotMoving* and *RejectAcquireIfMoving*) are analogous.[13]

**connector** *AllowMoveIfNotAcquiring*(*rflex.TSST*, *viam.AcS*)
**define** [*rflex.TSST*, *viam.AcS*]
**on** *rflex.TSST*, *viam.AcS*

**provided** $\neg$(*viam.AcS.active* $\wedge$
  *viamAcParams.bank.id* = "Marlin")
**do** {}

**connector** *RejectMoveIfAcquring*(*rflex.TSSR*, *viam.AcS*)
**define** [*rflex.TSSR*, *viam.AcS*]
**on** *rflex.TSSR*, *viam.AcS*
**provided** *viam.AcS.active* $\wedge$
  *viamAcParams.bank.id* = "Marlin"
**do** {*rflex.TSSR.rep* $\leftarrow$ CANNOT-ACQ-AND-MOVE}

Likewise, we have connectors to disallow taking pictures with the high resolution camera while the PTU is moving, and vice versa, and connectors also to disallow communication with an orbiter while moving, and vice versa, to ensure that communication is not disrupted. The connectors for the first constraint are similar to those shown above. The connectors for the second constraint are shown below. In what follows, *TSSA = TrackSpeedStartAbort*, *CAIS = CommunicateAbortIncompatibleServices*, *CT = CommunicateTrigger*, and *TSSS = TrackSpeedStartStatus*. As before, we only show the first two connectors; the other two are analogous.

**connector** *AllowCommIfNotMoving*(*antenna.CT*, *rflex.TSSS*)
**define** [*antenna.CT*, *rflex.TSSS*]
**on** *antenna.CT*, *rflex.TSSS*
**provided** $\neg$*rflex.TSSS.active*
**do** {}

**connector** *AbortMovingToComm*(*antenna.CAIS*, *rflex.TSSA*)
**define** [*antenna.CAIS'*, *rflex.TSSA*]
**on** *antenna.CAIS*, *rflex.TSSA*
**provided** *true*
**do** {*rflex.TSSA.rep* $\leftarrow$ CANNOT-COMM-AND-MOVE}
**on** *antenna.CAIS*
**provided** *true*
**do** {}

The following connectors prevent the total amount of power being drawn from the battery at any point in time from exceeding the power bus capacity.[14] This is done by rejecting new requests to start using the battery. In what follows, *SUB = StartUsingBattery*.

**connector** *AllowUseBatt*(*battery.SUBT*)
**define** [*battery.SUBT*]
**on** *battery.SUBT*
**provided** *FIDS.totalPwr* + *batterySUBParams.power* $\leq$
  *batteryInitParams.maxPwr*
**do** {}

**connector** *RejectUseBatt*(*battery.SUBR*)
**define** [*battery.SUBR*]
**on** *battery.SUBR*

---

[12]i.e., where the execution of the service returned a nominal report
[13]*Marlin* is the camera model, and *viamAcParams* is a variable that stores the camera model passed (as a parameter) with the most recent request to acquire an image.

[14]Recall that the Functional-IDS (FIDS) is a data structure that is accessible by all parts of a module and that stores data shared by the different parts of a module. Also, *batterySUBParams* is a variable that stores the power passed (as a parameter) with the most recent request to start using the battery.

**provided** *FIDS.totalPwr + batterySUBParams.power >
    batteryInitParams.maxPwr*
**do** {*battery.SUBR.rep ← MAX-PWR-EXCEEDED*}

Finally, the following connector prevents poster data produced by certain modules from being used if the data is not "fresh"; e.g., a speed reference produced by the NDD module is not "fresh" if it has not been updated for more than ten ticks. Recall that the *PosterAge* variable keeps track of the amount of time that has elapsed since the last time the associated *Poster* component was written to.

**connector** *AbtMoveIfPstrNotFresh(rflex.TSSA, ndd.RflexPoster.read)*
**define** [*rflex.TSSA, ndd.RflexPoster.read*]
**on** *rflex.TSSA, ndd.RflexPoster.read*
**provided** *ndd.RflexPoster.read.PosterAge > 10*
**do** {*rflex.TSSA.rep ← NDD-POSTER-NOT-FRESH*}

From the constraints presented in this section, it is clear that the BIP language provides a natural syntax for encoding non-trivial constraints on certain aspects of BIP components. Such constraints are fundamental for synthesizing a safe controller of the functional level. To make it even more convenient for the user, we have developed a higher-level language for specifying constraints;[15] these are then automatically translated into BIP connectors, such as those shown in this section. Examples of constraints in this new language are "$r_1$ before $r_2$," which guarantees that request $r_2$ is executed only after the successful execution of request $r_1$, and "$r_1$ does not overlap $r_2$," which guarantees that there is no point in time in which both $r_1$ and $r_2$ are executing.

## V. VERIFICATION WITH D-FINDER

Connectors like those shown in the previous section could cause deadlocks in the functional level, since they amount to adding tighter constraints to certain subsets of components. To check whether the additional connectors may cause deadlocks, and to determine whether (atomic and compound) components by themselves are free of deadlocks, we use D-Finder to first verify atomic components and to then incrementally verify the compound components resulting from their composition.

### A. Intra-module verification

We first discuss results from the verification we carried out for compound components corresponding to individual G<sup>en</sup><sub>o</sub>M modules. We start with a deadlock found while verifying the NDD (*Module*) component with D-Finder.

Figure 11 shows some of the components and associated connectors of the NDD component. Observe that there are three *Timer* components, one for the *Control Task* component, one for the *Execution Task* component, and one for the *Poster* component. A *Timer* consists of two ports. Port *tick* can be synchronized with other components (e.g., timers) to count the elapsed time before it reaches certain value. Port *trigger* will trigger some events when the elapsed time in terms of "ticks" reaches a certain amount, and reset the counter. To ensure that the duration between two contiguous *tick* synchronizations in a *Timer* component is equivalent to such a duration in any other *Timer* component, we strongly synchronize all *tick* ports of the mentioned *Timer* components with the *tick* port of the *MasterTimer* component. This component is the "global clock," which will effectively ensure that there are at least 10 milliseconds (ms) between two contiguous synchronizations involving all these *tick* ports.[16]

[15]This work is currently being published.

Although this design seemed correct, we found a non-trivial deadlock while verifying the NDD component with D-Finder. Intuitively, the reason for this deadlock is the strong synchronization between the *Timer* in *Control Task* and the *Timer* in *Execution Task*. Specifically, the deadlock scenario identified was the following: the Message Box is in location $abtI$; the *Scheduler* is in location $idle$; Execution Services *SetParams* and *GoTo* have started executing and they are respectively in locations $exec$ and $abrt$; variable $t$ in the *Timer* of the *Execution Task* (*ExecTaskTimer*) has been reset to zero; and variable $t$ in the *Timer* of the *Control Task* (*InterfaceTimer*) has reached the maximal value. Recall that, like all *Timer* components, the *InterfaceTimer* simply makes the *trigger* port available when the "ticks" reach a predefined value.

The deadlock was caused by a cyclic-waiting of events (interactions) between two timers and the related components. Roughly speaking, *InterfaceTimer* waits for *Message Box*; *Message Box* waits for some event indirectly controlled by *Scheduler*; *Scheduler* waits for the trigger from *ExecTaskTimer*; and *ExecTaskTimer* waits for *InterfaceTimer*'s trigger to continue the "tick."

More precisely, in this scenario, *Scheduler* is waiting to synchronize with the $trigger$ port of *ExecTaskTimer* in order to start the next round of execution; *ExecTaskTimer* is waiting for variable $t$ in *InterfaceTimer* to be reset (via the synchronization involving its $trigger$ port) in order to continue with the synchronization between the four connected $tick$ ports; *InterfaceTimer* is waiting for *Message Box* to return to location $idle$ via location $give$, so that the *InterfaceTimer* can reset its variable $t$ and perform the synchronization with the four connected $tick$ ports; and *Message Box*, after having aborted the *GoTo* service, is waiting to trigger the *Stop* service, in order to return to location $idle$ via location $give$ (see Figure 8). However, it is not possible to trigger the *Stop* service because, according to our mapping from G<sup>en</sup><sub>o</sub>M to BIP, no condition on the transition corresponding to any $trig_{b_i}$ port in the *Message Box* component will be met. Specifically, this is because the *SetParams* and *GoTo* services have already started executing, and all other services in the NDD module have been declared by the user as being incompatible with at least one of these services. Consequently, a deadlock state has been reached.

Since the deadlock was caused by the cyclic-waiting between the two timers and the related components, we had to break the cycle. Our solution was to modify the connector synchronizing the *tick* ports to allow *InterfaceTimer* to not participate in the synchronization via the connector if it cannot participate. In precise terms, we have replaced the strong synchronization between the *Timer* components in Figure 11 with two new connectors, of which the first is shown below.

[16]More than 10 ms may be taken if at least one of the *Timer* components takes time to complete their *trigger* synchronizations. Waiting for 10 ms is implemented using the *usleep* system call. There is ongoing work [39] to extend BIP with the ability to model time, which will remove the need for the system call and improve the accuracy of measuring time.

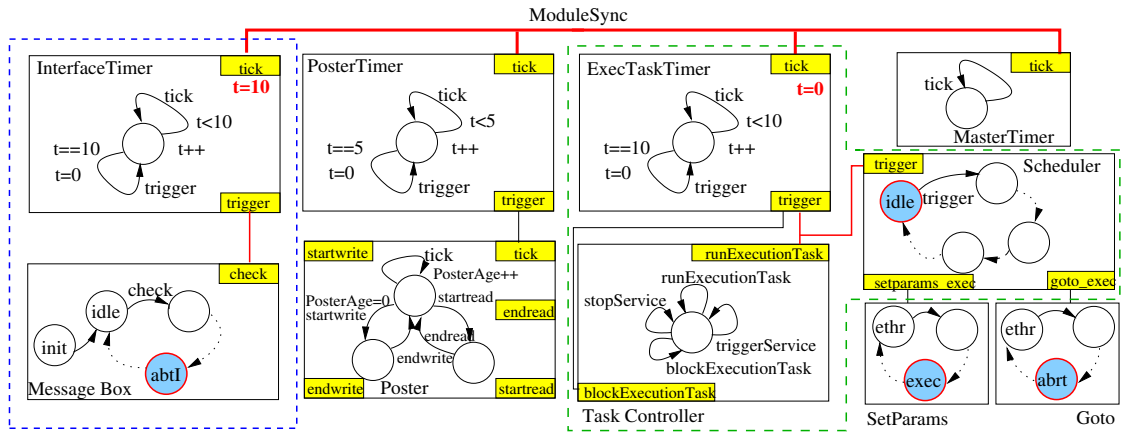

Fig. 11. A deadlock scenario caused by synchronization between *Timer* components. Dotted lines stand for ignored locations and transitions.

**connector** *ModuleSync*(*execTaskTimer.tick*, *posterTimer.tick*, *interfaceTimer.tick*)
**define** [*execTaskTimer.tick*, *posterTimer.tick*]′,
 *interfaceTimer.tick*
**export** port Port *moduleTick*

The above connector, exported as *moduleTick*, executes a strong synchronization between the three *tick* ports if the *tick* port of the *InterfaceTimer* component is available, and otherwise, the connector executes a strong synchronization only between the *tick* ports of the *ExecTaskTimer* component and *PosterTimer* component. This solves the deadlock because it allows the *Timer* components inside the *Module* component to continue executing even if the *Message Box* component is waiting for a service to be aborted. The second connector is shown below.

**connector** *InterModuleSync* (*masterTimer.tick*, *moduleTick*$_1$,
 . . . , *moduleTick*$_n$)
**define** *masterTimer.tick*, *moduleTick*$_1$, . . . , *moduleTick*$_n$
**on** *masterTimer.tick*, *moduleTick*$_1$, . . . , *moduleTick*$_n$
**provided** *true*
**do** {}

The above connector is for global synchronization between all *Module* components contained in the functional level, where $\{moduleTick_i\}_{1 \leq i \leq n}$ is a set of connectors of type *ModuleSync*, one for each *Module* component in the functional level composed of $n$ *Module* components.

*B. Inter-module verification*

Even after correcting individual modules with respect to deadlocks, it is still possible for collections of modules to contain deadlocks or to exhibit unsafe behavior. In this section, we discuss some of the more interesting results from our analysis of groups of modules for deadlock-freedom and "data freshness."

Consider once again the two connectors described above, which synchronized all the *Module* components in the functional level by synchronizing a single *MasterTimer* component with the associated *Timer* components of all *Control Task*, *Execution Task*, and *Poster* components. As one might expect, with such a complex synchronization there may be the risk of a deadlock if any of the *Timer* components cannot perform a *tick* transition due to its *trigger* port never becoming available after the *Timer*'s period is reached. Interestingly, we successfully verified using D-Finder that the synchronization between the *Timer* components belonging to `NDD`, `Aspect` and `LaserRF` modules are deadlock-free, and moreover, that the synchronization between the *Timer* components of `NDD` and `RFLEX` are also deadlock-free.

Finally, we verified the "data freshness" property discussed in Section IV-B. Recall that this property requires data in posters to be below a certain user-specified age, given in terms of "ticks." Figure 12 shows the components and associated connectors between the `Aspect` module (components and connectors within the left dashed box) and the `NDD` module (components and connectors within the right dashed box). Observe the two connectors *InterModuleSync* and *PermStartRead*, which respectively synchronizes the *Timer* components belonging to certain different modules, and connects ports of component *PolarPoster* in module `Aspect` and those of components *Permanent*, *Scheduler* and *Lock* in module `NDD`, in order to allow `NDD` to take into account the *PosterAge* variable of *PolarPoster*. Observe also the constraint $PosterAge < 5$ on connector *PermStartRead*, which prevents `NDD` from reading the data stored in *PolarPoster* if the data is older than or equal to 5.[17]

---

[17]Observe from Figure 12 that for every increment of *PosterAge* in *PolarPoster* there are five additional ticks in *PosterTimer*.

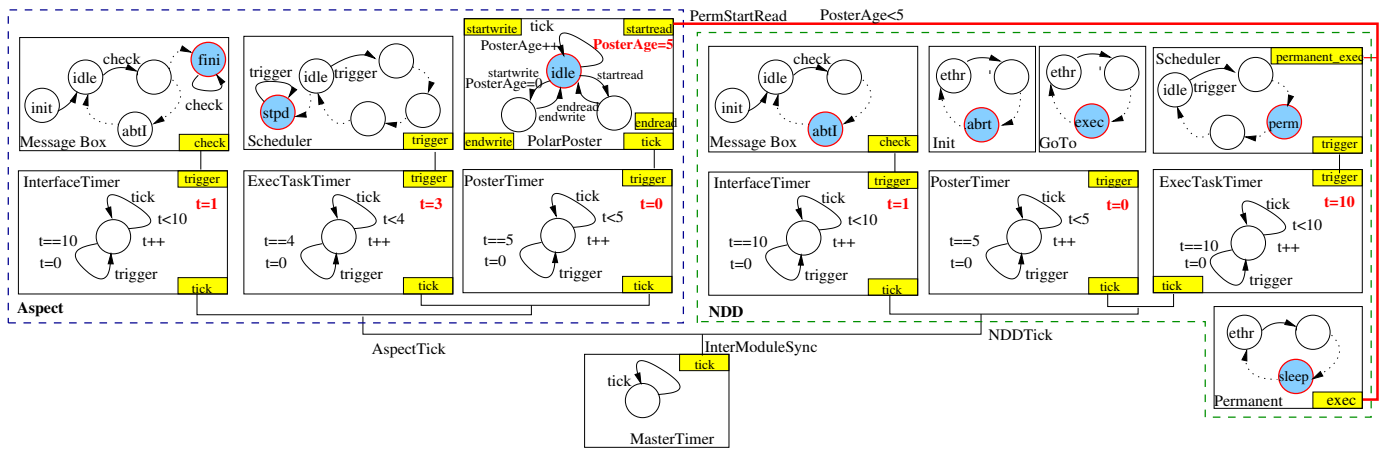

Fig. 12.  Deadlock scenario with data-freshness constraint.

In this design, D-Finder detected a deadlock caused by the *PermStartRead* connector. This deadlock is illustrated in Figure 12. Specifically, the deadlock scenario identified was the following.[18] For the `Aspect` module, the *Message Box* component is in location *fini*, the *Scheduler* component is in location *stpd*; the *PolarPoster* component is in location *idle* with $PosterAge = 5$; variable $t$ in the *PosterTimer* component is reset to zero; and variable $t$ in the *ExecTaskTimer* component is 3. For the `NDD` module, the *Message Box* component is in location *abtI*; the *Scheduler* component is in location *perm*; the *Permanent* component is in location *sleep*; the *GoTo* and *Init* components are respectively in locations *exec* and *abrt*; variable $t$ in *PosterTimer* is reset to zero; and variable $t$ in *ExecTaskTimer* is 10.

Intuitively, the reason for the deadlock is as follows. When $PosterAge = 5$, $PermStartRead$ cannot be executed causing components *Scheduler* and *Permanent* to be blocked. Because *Scheduler* interacts with its timer and other execution services, the related components and *Message Box* in module `NDD`, as well as the *MasterTimer* component are blocked. Consequently, module `Aspect` is also blocked.

The detailed reason for this deadlock is as follows. First, since $PosterAge = 5$ holds, the constraint on $PermStartRead$ disallows an interaction with this connector, resulting in *Scheduler* and *Permanent* components being blocked. Consequently, since $t = 10$ and (due to *Scheduler* being blocked) an interaction is not possible with the *trigger* (and hence *tick*) port of the `NDD` module's *ExecTaskTimer* component, an interaction is also not possible via the *InterModuleSync* connector. This entails that all the *Timer* components of the `Aspect` module are blocked. Moreover, since the `Aspect` module has stopped after completing the execution of its services, the whole module is blocked. Finally, since both *Init* and *GoTo* components are executing, no condition on the transition from location *abtI* in the *Message Box* component of `NDD` will be met (for reasons explained previously in the first deadlock's discussion), resulting in the *Message Box* having to wait for the termination of *Execution Service Init* or *GoTo*, which are both blocked due to the blocked *Scheduler*. Consequently, a (global) deadlock state has been reached.

We tried to resolve this deadlock by adding a connector to abort the *GoTo* component of `NDD` whenever the data of the `Aspect` module is not fresh, as follows:

**connector** *GoToAbortNonFreshData* (*Scheduler.permanentexec*, *Permanent.exec*, *PolarPoster.read*, *NDDLock.take*, *GoTo.abort*)
**define** *Scheduler.permanentexec*, *Permanent.exec*,
  *PolarPoster.read*, *NDDLock.take*, *GoTo.abort*
**on** *Scheduler.permanentexec*, *Permanent.exec*, *PolarPoster.read*,
  *NDDLock.take*, *GoTo.abort*
**provided** *PolarPoster.read.PosterAge* $\geq 5$
**do** {}

However, we still obtained via D-Finder a deadlock scenario similar to the above. The only difference was that *GoTo* was in location *abrt* instead of in location *exec*, from where port *abort* is not available. The solution to this deadlock was to abort *GoTo* only in situations where *(i)* `NDD` is accessing the non-fresh data from *PolarPoster*, and *(ii)* *GoTo* is actually executing. The modified connector is shown below.

**connector** *GoToAbortNonFreshData* (*Scheduler.permanentexec*, *Permanent.exec*, *PolarPoster.read*, *NDDLock.take*, *GoTo.abort*)
**define** [*Scheduler.permanentexec*, *Permanent.exec*,
  *PolarPoster.read*, *NDDLock.take*]′, *GoTo.abort*
**on** *Scheduler.permanentexec*, *Permanent.exec*, *PolarPoster.read*,
  *NDDLock.take*, *GoTo.abort*
**provided** *PolarPoster.read.PosterAge* $\geq 5$
**do** {}
**on** *Scheduler.permanentexec*, *Permanent.exec*, *PolarPoster.read*,
  *NDDLock.take*
**provided** *PolarPoster.read.PosterAge* $\geq 5$
**do** {}

Note that unlike the deadlock illustrated in Figure 11, the deadlock between the `NDD` and `Aspect` modules is not caused solely by connectors involving *tick* ports: it is caused by an overly-restrictive condition on an interaction. The fact that the interaction is not possible also blocks the execution of other interactions, including those involving *tick* ports. Hence, the deadlock would still be present even if we remove the explicit timing connectors and, for instance, use real-time "clocks" in components such as *Message Box* and *Scheduler*, which we have done in more recent work.[19]

---

[18]Note that the components' locations in Figure 12 could be reached if *Init* is triggered first, *GoTo* is triggered after the successful completion of *Init*, and then *Init* is again triggered soon after *GoTo* but aborted almost immediately.

## VI. Experimental results

The time taken to generate a BIP module from a G$^{en}_o$M module is negligible. Table I shows the time taken for computing invariants for the deadlock-freedom checking of thirteen modules and a group of modules by D-Finder. *Module* stands for names of modules; *Component* stands for the total number of atomic components in a module; *Locations* stands for the total number of control locations in a module; *Interactions* stands for the total number of interactions in a module; *States* stands for the total number of states in a module—including those in its constituent components; *LOC* stands for the number of lines of (BIP) code in a module; and *Minutes* stands for the time taken by D-Finder to return a result. Observe from the table that we were able to check for the deadlock-freedom of all our (single) modules in reasonable amounts of time, even for those consisting of thousands of lines of BIP code. This shows that D-Finder can be used to verify complex, real-world domains, and not just toy examples as shown in previous work [40]. It was already shown in [40], [34], [35] that the component sizes handled by D-Finder are far beyond those that can be handled by other state of the art academic verification tools such as NuSMV [41] and SPIN [42]. Initially, the deadlock-freedom checking of certain groups of modules (namely, the three modules LaserRF, Aspect and NDD, and the two modules NDD and RFLEX) took over 40 minutes, which was expected considering the large state spaces of the two groups. However, with a recently improved version of the D-Finder tool [43], we were able to confirm within a matter of a few minutes that the synchronization between all the five related *Module* components shown in the table were deadlock-free.

The runtime performance of BIP and G$^{en}_o$M is shown in Table II. The results shown are the CPU time taken by each module, obtained via the *TIME+* column of the *top* unix command. The results are an average over five runs, on a machine with 1 GB of RAM, and two Intel Pentium 4, 3.06 GHz CPUs with 512 KB of cache on each. To have a fair comparison between G$^{en}_o$M and BIP, no safety constraints were included in the BIP model, and no red rocks (objects) were in the vicinity.[20] The experiment conducted was simpler than the one introduced in Section IV in that the rover only explored a single location, rather than multiple locations, before returning back to the initial location. For a more complex and complete experiment, we invite the reader to check the video online: http://db.tt/H4oRLva (please download it to see it in full resolution and with the proper captions). It shows a complete experiment with the Dala rover running the functional level (see Figure 1) in BIP, with various fault injections leading the BIP engine to prevent them and report them to the decisional level (OpenPRS in this case), which takes corrective actions.

Independent samples t-tests confirmed that the differences between the values in the two columns of Table II are statistically significant, with P-values of 0.00065 or lower.[21] From the results, we can see that the total time taken by the BIP modules is 27927 time units, whereas the total time taken by the G$^{en}_o$M modules is 2731 time units. Hence, BIP took approximately ten times more CPU time than G$^{en}_o$M. Likewise, in terms of the usage of the CPUs immediately before the end of the experiments, the G$^{en}_o$M experiment used approximately ten times less than the BIP experiment, with an average usage of 6.3% for the former and 52% for the latter. This additional overhead with BIP comes from the decision making by the BIP engine. In particular, the BIP engine actively computes all the feasible interactions at every cycle, where the time taken for this computation is proportional to the number of interactions in the BIP modules. This explains why the RFLEX and VIAM modules are two of the slowest and the Antenna and Heating modules are two of the fastest. The number of interactions, however, is not the only factor that affects the speed of modules. Currently, the BIP engine runs continuously without halting, even when there are no interactions to fire. As a result, it uses all the CPU it can get; simply forcing the BIP engine's main loop to run at 25 Hz divides the CPU load by two. Another possible improvement is to be smarter about the guards which are evaluated at each loop, e.g., to only check those in which the associated variable values have changed. As discussed in Section II-E, from the feasible interactions computed, priority rules are applied to select one interaction, the data transfer of the interaction is performed, and transitions on atomic components are executed.

Despite the CPU overhead incurred by using BIP, the time taken to run the complete experiment was approximately four and half minutes on average in both G$^{en}_o$M and BIP, with BIP taking about four seconds longer on average. Therefore, the overhead in using BIP was negligible when taking into account the total time for executing the mission. This is most likely because in the BIP experiment the CPUs are used at a much higher capacity (52% usage) than in the G$^{en}_o$M experiment (6.3% usage), which would compensate for the extra CPU time needed by the BIP modules. The additional CPU time required by BIP was also perhaps mitigated by the time taken up in executing actions in the real world, such as moving the robot, communicating, and moving the PTU. In our experiments, simply moving Dala from $(x, y)$ coordinates $(0, 0)$ to $(4, 0)$ takes approximately 30 seconds, moving the PTU toward the left or right front wheels of the rover takes approximately 5 seconds, and transmitting a single picture to the orbiter also takes approximately 5 seconds.

---

[19] In fact, we also found another (albeit trivial) deadlock in the *Antenna* module that was not related to *tick* ports, but had to do with obtaining a "lock" for a component (via the *IDS Lock* component) but then never releasing that "lock."

[20] Note that the Hueblob module was still used to continuously search for red rocks.

[21] Specifically, the P-values for the two (five-member) sets of runtime results from the top row (LaserRF) to the bottom row (VIAM) are respectively as follows: $2.31 \times 10^{-6}, 2.07 \times 10^{-11}, 2.00 \times 10^{-7}, 1.83 \times 10^{-11}, 6.53 \times 10^{-4}, 4.23 \times 10^{-10}, 7.16 \times 10^{-8}, 4.27 \times 10^{-10}, 8.07 \times 10^{-7}$, and $1.09 \times 10^{-11}$.

TABLE I
Results for deadlock-freedom checking.

| Module | Components | Locations | Interactions | States | LOC | Minutes |
|---|---|---|---|---|---|---|
| LaserRF | 43 | 213 | 202 | $2^{20} \times 3^{29} \times 34$ | 4353 | 1:22 |
| Aspect | 29 | 160 | 117 | $2^{17} \times 3^{23}$ | 3029 | 0:39 |
| NDD | 27 | 152 | 117 | $2^{22} \times 3^{14} \times 5$ | 4013 | 8:16 |
| RFLEX | 56 | 308 | 227 | $2^{34} \times 3^{35} \times 1045$ | 8244 | 9:39 |
| Antenna | 20 | 97 | 73 | $2^{12} \times 3^{9} \times 13$ | 1645 | 0:14 |
| Battery | 30 | 176 | 138 | $2^{22} \times 3^{17} \times 5$ | 3898 | 0:26 |
| Heating | 26 | 149 | 116 | $2^{17} \times 3^{14} \times 145$ | 2453 | 0:17 |
| PTU | 37 | 174 | 151 | $2^{19} \times 3^{22} \times 35$ | 8669 | 0:59 |
| Hueblob | 28 | 187 | 156 | $2^{12} \times 3^{10} \times 35$ | 3170 | 5:42 |
| VIAM | 41 | 227 | 231 | $2^{10} \times 3^{6} \times 665$ | 5099 | 4:14 |
| DTM | 34 | 198 | 201 | $2^{28} \times 3^{20} \times 95$ | 4160 | 13:42 |
| Stereo | 33 | 196 | 199 | $2^{27} \times 3^{20} \times 95$ | 3591 | 13:20 |
| P3D | 50 | 254 | 219 | $2^{13} \times 3^{5} \times 5^{4} \times 629$ | 6322 | 3:51 |
| LaserRF+ Aspect + NDD + RFLEX + Antenna | 171 | 926 | 732 | $2.9 \times 10^{90}$ | 21284 | 5:05 |

TABLE II
G$^{en}$₀M and BIP runtime performance. A time unit is one hundredth of a CPU second.

| Module | G$^{en}$₀M execution time | BIP execution time |
|---|---|---|
| LaserRF | 120 | 1947 |
| Aspect | 192 | 2362 |
| NDD | 43 | 2009 |
| RFLEX | 168 | 10763 |
| Antenna | 56 | 1102 |
| Battery | 69 | 1219 |
| Heating | 92 | 1029 |
| PTU | 126 | 2394 |
| Hueblob | 690 | 1850 |
| VIAM | 1046 | 3225 |

Thus, this experiment indicates that the code generated from BIP models can be run on realistic robotic platforms such as Dala, even those with only modest processing power and memory capacities. While we acknowledge that certain—perhaps smaller—robotic platforms with more stringent processing power and memory capacities may not be able to handle the additional CPU overhead in using BIP, we also note that many of the newer robots have multi-core or many-core CPUs, and much higher memory capacities than Dala. Moreover, with the possible improvements to BIP already mentioned, we expect code corresponding to BIP models to have a significantly lower overhead on robotic hardware.

## VII. Conclusion

Despite the fact that software has become a large part of robot development, one must admit that the software models used up to now are either too coarse, too high level, or too large and thus very difficult to analyze and verify. We propose a novel approach for developing functional levels of robotic systems, which incorporates a component-based design approach (BIP) in an existing architectural tool for developing functional modules (G$^{en}$₀M). Our approach allows the synthesis of a controller that is correct by construction that encodes and enforces designer-supplied safety properties. Moreover, we use the D-Finder tool to formally verify that a significant part of our functional level is deadlock-free, and that it conforms to other safety properties such as data freshness.

We were able to run experiments with a complete functional and decisional level on the Dala rover, and to demonstrate via fault injections that the BIP engine successfully stops the robot from reaching undesired/unsafe situations like those discussed previously, and that it reports faults appropriately to the decisional level. In terms of runtime performance, experiments show that by using our integrated BIP framework, instead of using G$^{en}$₀M alone, the CPU time taken is approximately ten times more than that taken by G$^{en}$₀M alone, mainly because the BIP engine must compute all the feasible interactions at each step in its execution. Nonetheless, our robotic platform was able to handle this additional overhead, and the overall time taken for the mission was not affected by it.

We plan to extend our work in various directions: e.g., a real-time BIP engine to take into account wall clock properties, and a distributed engine to distribute it over more than one CPU. Another more ambitious research direction is to study the use of the BIP approach at the decisional level of our autonomous robot.

## Acknowledgments

The authors would like to thank Matthieu Gallien for the implementation of the G$^{en}$₀M-to-BIP tool and for other useful discussions. We also thank the other participants of the GOAC and MARAE projects for their inputs.